\documentclass[runningheads]{llncs}

\usepackage{eccv}

\usepackage{eccvabbrv}

\usepackage{graphicx}
\usepackage{booktabs}
\usepackage[misc]{ifsym}
\newcommand{\task}{Continual Missing Modality Learning}
\newcommand{\shorttask}{CMML}
\newcommand{\method}{Reconstruct before Query}
\newcommand{\shortmethod}{RebQ}
\newcommand{\datasetone}{UPMC-Food101-CMML}
\newcommand{\datasettwo}{MM-IMDb-CMML}
\usepackage{booktabs}
\usepackage{multirow}

\usepackage{algorithm}
\usepackage{algpseudocode}

\usepackage{xcolor}
\definecolor{myblue}{RGB}{33,110,233}
\definecolor{myorange}{RGB}{250,104,0}
\definecolor{mypurple}{RGB}{181,33,255}
\definecolor{mypink}{rgb}{1.0, 0.08, 0.58}
\definecolor{mygreen}{RGB}{0, 180, 0}

\usepackage[accsupp]{axessibility}  

\usepackage{hyperref}

\usepackage{orcidlink}

\begin{document}

\title{Reconstruct before Query: Continual Missing Modality Learning with Decomposed Prompt Collaboration} 

\titlerunning{RebQ: Reconstruct before Query}

\author{Shu Zhao\inst{1} \and
Xiaohan Zou\inst{1} \and
Tan Yu\inst{2} \and
Huijuan Xu\inst{1}}

\authorrunning{RebQ: Reconstruct before Query}

\institute{Pennsylvania State University \and NVIDIA\\
\email{smz5505@psu.edu}}

\maketitle

\begin{abstract}
    
    Pre-trained large multi-modal models~(LMMs) exploit fine-tuning to adapt diverse user applications. Nevertheless, fine-tuning may face challenges due to deactivated sensors~(e.g., cameras turned off for privacy or technical issues),  yielding modality-incomplete data and leading to inconsistency in training data and the data for inference. Additionally, continuous training leads to catastrophic forgetting, diluting the knowledge in pre-trained LMMs. To overcome these challenges, we introduce a novel task, {\task}~({\shorttask}), to investigate how models can generalize when data of certain modalities is missing during continual fine-tuning. Our preliminary benchmarks reveal that existing methods suffer from a significant performance drop in {\shorttask}, even with the aid of advanced continual learning techniques. Therefore, we devise a framework termed \textbf{Re}construct \textbf{b}efore \textbf{Q}uery~({\shortmethod}). It decomposes prompts into modality-specific ones and breaks them into components stored in pools accessible via a key-query mechanism, which facilitates Parameter-Efficient Fine-Tuning\footnote{In this paper, fine-tuning denotes parameter-efficient fine-tuning unless specified otherwise.} and enhances knowledge transferability for subsequent tasks. Meanwhile, our {\shortmethod} leverages extensive multi-modal knowledge from pre-trained LMMs to reconstruct the data of missing modality. Comprehensive experiments  demonstrate that {\shortmethod}  effectively reconstructs the missing modality information and retains pre-trained knowledge. Specifically, compared with the baseline,  RebQ improves average precision from $20.00$ to $50.92$  and decreases  average forgetting  from $75.95$ to $8.56$. Code and datasets are available on \url{https://github.com/Tree-Shu-Zhao/RebQ.pytorch}
    
\end{abstract}
\begin{figure}[t] \label{fig:task}
\begin{center}
    \includegraphics[width=0.85\linewidth]{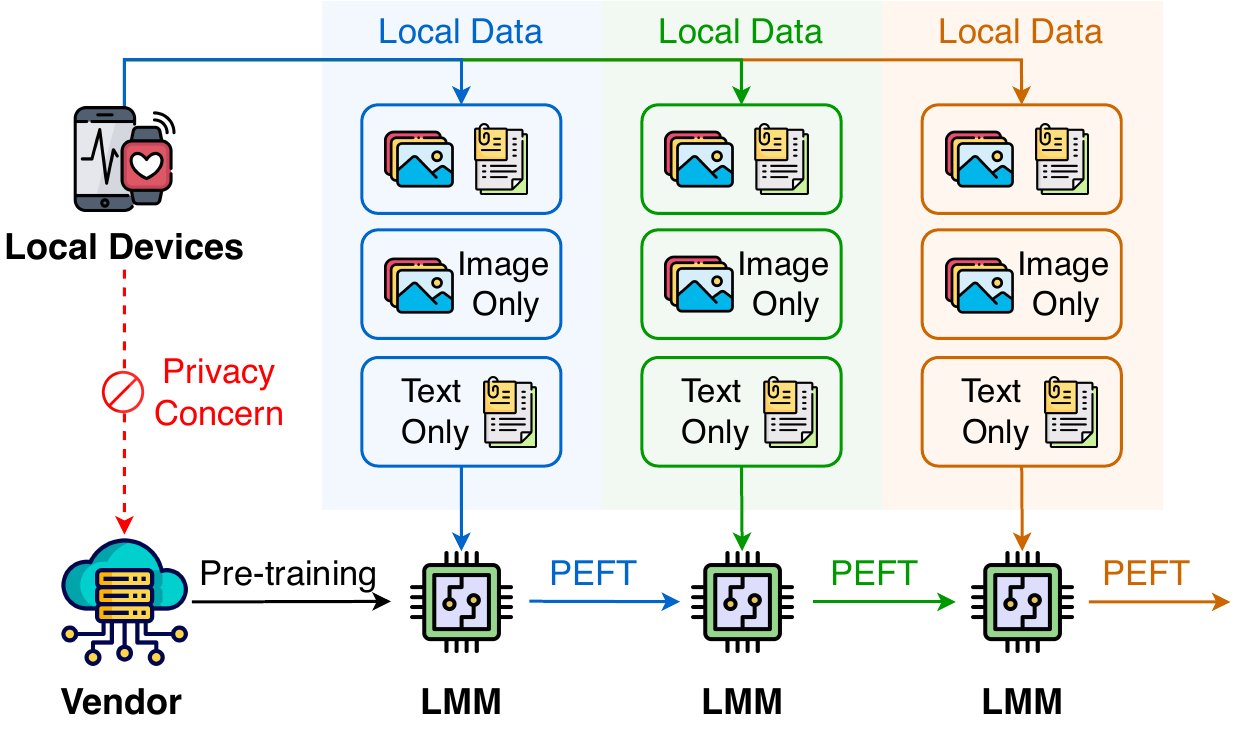}
\end{center}
\caption{Large Multi-Modal Models~(LMMs) pre-trained on a huge amount of multi-modal data often necessitate subsequent Parameter-Efficient Fine-Tuning~(PEFT) to adapt to diverse user applications.
However, deactivated sensors~(e.g., cameras turned off for privacy or technical issues) yield modality-incomplete data, significantly challenging the fine-tuning process.}
\label{fig:task}
\end{figure}

\section{Introduction}

Large multi-modal Models (LMMs), pre-trained in cloud-based computing platforms with abundant resources, have powerful understanding and generating capability for multi-modal data~\cite{liu2023improvedllava,liu2023llava,chen2023minigptv2,zhu2023minigpt,DBLP:journals/corr/abs-2309-10020,DBLP:journals/corr/abs-2311-05437,DBLP:journals/corr/abs-2311-00571,DBLP:journals/corr/abs-2310-15166,DBLP:journals/corr/abs-2310-07704,DBLP:journals/corr/abs-2309-17102,DBLP:journals/corr/abs-2305-14725,DBLP:journals/corr/abs-2303-13519,DBLP:conf/cvpr/FuYZFSWB23,DBLP:journals/corr/abs-2210-10841,DBLP:conf/cikm/YuLJYFL22}. To adapt to diverse user environments, LMMs often necessitate subsequent fine-tuning with user data. Normally, it is beneficial to continue the learning process over the device's lifespan~\cite{DBLP:journals/corr/abs-2306-03310,DBLP:journals/corr/abs-2307-06870}, as illustrated in Figure~\ref{fig:task}. However, due to privacy concerns~\cite{DBLP:journals/db/HarborthP20}, users may resist sending their data back to the model vendors for fine-tuning. Consequently, model adaptation should be iteratively conducted on user devices with limited resources, obviating the transmission of sensitive data externally. Moreover, deploying all modalities in practical applications might be unfeasible due to security concerns or technical issues~\cite{DBLP:conf/aaai/WooLPNK23,DBLP:conf/cvpr/LeeTCL23,DBLP:conf/cvpr/0005CMAH023,DBLP:journals/corr/abs-2306-1279,DBLP:conf/cvpr/GongMDBLWR23}. For instance, consider a long-term surveillance system that continuously monitors an environment using multi-modal sensors (e.g., cameras, microphones, and thermal sensors). Over time, some sensors may be temporarily unavailable, leading to missing modalities. Additionally, the system needs to adapt to changing conditions, such as varying lighting, weather, and the introduction of new objects or activities. Accordingly, LMMs must be robust and adaptive, capable of operating in dynamic and uncertain environments.

In this work, we define a novel task termed {\task}~({\shorttask}) and construct two datasets, {\datasetone} and {\datasettwo}, to benchmark the challenging settings in real-world scenarios. This paradigm evaluates models in handling non-static data with absent modalities. We reveal that existing methods for the missing modality problem are susceptible to severe performance degradation due to catastrophic forgetting~\cite{DBLP:conf/nips/French93}. Even combined with continual learning strategies, these methods struggle to capture knowledge across modalities and fail to yield satisfactory outcomes. 

Inspired by human learning~\cite{alvarez1994memory,ratcliff1990connectionist}, we devise a novel method named {\method}~({\shortmethod}). {\shortmethod} utilizes a pre-trained LMM with extensive multi-modal knowledge against the modality-incomplete data~\cite{DBLP:conf/cvpr/0002R0T022,DBLP:conf/cvpr/LeeTCL23}, which freezes the parameters within the pre-trained LMM and leverages prompts to facilitate parameter-efficient fine-tuning. Specifically, it decomposes unified prompts into modality-specific ones and then further breaks them into components stored in pools accessible via a key-query mechanism, significantly enhancing knowledge transferability for subsequent tasks. However, accessing pools requires corresponding modality-specific queries, which is not feasible when the modality-incomplete data comes. Therefore, {\shortmethod} explicitly harnesses the extensive multi-modal knowledge from the pre-trained LMM to reconstruct the absent query for accessing the relevant components.

The main contributions are outlined as follows:
\begin{itemize}
    \item To the best of our knowledge, this paper is the first to introduce the challenging {\task} task for non-static modality-incomplete data. This setting is commonly encountered in real applications but rarely exploited in academia.
    \item We construct two datasets to benchmark the {\shorttask} task, showing that baselines suffer from significant performance challenges for missing modality techniques, even when combined with continual learning approaches.
    \item We propose {\method}~(\shortmethod) to decompose prompts into modality-specific components and explicitly harness the multi-modal knowledge from a pre-trained model to reconstruct the absent query for accessing the relevant components.
    
\end{itemize}
 
\section{Related Work}

\subsection{Missing Modality}
The issue of missing modalities presents a significant challenge in computer vision~\cite{DBLP:conf/ijcai/ZhaoWZL021,DBLP:conf/mm/ZhaoWZ00020,DBLP:journals/corr/abs-2310-01356,DBLP:journals/corr/abs-2310-01358,liu2023improvedllava,liu2023llava,chen2023minigptv2,zhu2023minigpt,DBLP:journals/corr/abs-2309-10020,DBLP:journals/corr/abs-2311-05437,DBLP:journals/corr/abs-2311-00571,DBLP:journals/corr/abs-2310-15166,DBLP:journals/corr/abs-2310-07704,DBLP:journals/corr/abs-2309-17102,DBLP:journals/corr/abs-2305-14725,DBLP:journals/corr/abs-2303-13519,DBLP:conf/cvpr/FuYZFSWB23,DBLP:journals/corr/abs-2210-10841,DBLP:conf/cikm/YuLJYFL22}, particularly when certain data modalities, integral during training, become inaccessible at the inference stage. A breadth of studies~\cite{DBLP:conf/cvpr/HoffmanGD16,DBLP:conf/eccv/LuoHJNF18,DBLP:conf/eccv/GarciaMM18,DBLP:journals/pami/GarciaMM20,DBLP:conf/aaai/MaRZTWP21,DBLP:conf/cvpr/0002R0T022,DBLP:conf/aaai/WooLPNK23,DBLP:conf/cvpr/LeeTCL23,DBLP:conf/cvpr/ShvetsovaCR0KFH22,DBLP:conf/cvpr/0005CMAH023,DBLP:journals/corr/abs-2306-1279,DBLP:conf/cvpr/GongMDBLWR23} has ventured into resolving this problem. Some research endeavors~\cite{DBLP:conf/cvpr/HoffmanGD16,DBLP:conf/eccv/LuoHJNF18,DBLP:conf/eccv/GarciaMM18,DBLP:journals/pami/GarciaMM20,DBLP:conf/aaai/MaRZTWP21} have harnessed auxiliary modalities' side information during training, empowering models to extrapolate the absent modality information in the inference phase. This technique notably improves the model's effectiveness in missing modality learning. Conversely, alternative methodologies~\cite{DBLP:conf/cvpr/ShvetsovaCR0KFH22,DBLP:conf/cvpr/0005CMAH023,DBLP:journals/corr/abs-2306-1279,DBLP:conf/cvpr/GongMDBLWR23} strive to align embedding spaces across diverse modalities, thereby allowing incomplete modalities to ascertain their coordinates within a unified embedding framework during inference. 

A recent and notable shift in the landscape has been the adoption of utilizing tokens within pre-trained transformer architectures to represent the information of missing modalities~\cite{DBLP:conf/cvpr/0002R0T022,DBLP:conf/aaai/WooLPNK23,DBLP:conf/cvpr/LeeTCL23,DBLP:journals/corr/abs-2312-15890}. This innovative approach has gained considerable interest due to its ability to effectively encapsulate modality information with limited resources. However, these approaches are trained and tested on fixed datasets,  struggling with the non-static data common in real-world applications. In contrast, in the work, we target to solve the missing-modality issue in non-static settings. We incorporate prompt tokens to invoke the intrinsic knowledge within pre-trained models, subsequently rendering them into concise surrogates for the missing modalities, which not only streamlines the representation of absent information but also capitalizes on knowledge in models, thereby enhancing the robustness and adaptability of models in scenarios confronted with missing modalities.

\subsection{Continual Learning}
Continual learning is dedicated to the enhancement of a model's knowledge through an ongoing series of tasks, crucially maintaining the integrity of information acquired in earlier stages. This field can be broadly segmented into four strategies: regularization-based, rehearsal-based, architecture-based, and prompt-based approaches.

Regularization-based methods~\cite{DBLP:conf/eccv/AljundiBERT18,DBLP:conf/eccv/LiH16,DBLP:conf/icml/ZenkePG17, dong2023heterogeneous, wang2023task, goswami2023attribution, lin2022beyond, oh2022alife} introduce constraints during the training on subsequent tasks, ensuring the modification of learned weights remains less divergent from established patterns. While preserving previous knowledge, these methods can inadvertently restrain the model, potentially undermining performance advancements due to over-emphasis on stability.

Rehearsal-based methods~\cite{DBLP:conf/nips/ShinLKK17,DBLP:conf/icml/GaoL23a,DBLP:conf/nips/Lopez-PazR17,DBLP:conf/eccv/PrabhuTD20, DBLP:conf/icml/JeeveswaranBZA23, liu2023augmented, sun2023decoupling, sun2023regularizing, cha2023rebalancing, zhu2023continual, lin2023pcr, liu2023continual, luo2023class, DBLP:conf/iclr/0001WYZ23, DBLP:conf/iclr/BhatZA23, DBLP:conf/iclr/SarfrazAZ23, sarfraz2023sparse, zhang2022simple, sun2022exploring} advocate for retaining a memory buffer composed of exemplars from previous tasks to replay them in the current task. The efficacy of such methods is inherently tied to the capacity of the memory buffer.  Nevertheless, privacy concerns may preclude the retention of specific samples, imposing limitations on their feasibility.

Architecture-based methods devise distinct components for new tasks, manifesting as network expansion~\cite{DBLP:conf/icml/LiZWSX19,DBLP:conf/nips/RaoVRPTH19, ge2023clr, rymarczyk2023icicle, wang2023task, hu2023dense, wang2022beef, DBLP:conf/iclr/BhatZA23, jiang2023neural, ermis2022memory} or the formulation of task-dedicated sub-networks~\cite{DBLP:conf/nips/KeLH20,DBLP:conf/cvpr/MallyaL18,DBLP:conf/icml/SerraSMK18, DBLP:conf/icml/KonishiKOKK023, xiao2023endpoints}. Despite achieving satisfactory task-specific precision, these methods need to predicate the availability of task identity during the testing phase, which is a condition not guaranteed in more dynamic and realistic  task-agnostic environments.

Prompt-based methods~\cite{DBLP:conf/cvpr/0002ZL0SRSPDP22,DBLP:conf/nips/WangHH22,DBLP:conf/eccv/0002ZESZLRSPDP22,DBLP:conf/cvpr/SmithKGCKAPFK23,DBLP:conf/cvpr/VillaAAAHHSG23,jung2023generating,khan2023introducing} emerge as marks of a significant evolution in continual learning paradigms. They ingeniously utilize learnable prompts for each task, extracting and applying knowledge from pre-trained models. This strategy is not only highly efficient in terms of parameter usage but also circumvents the need for extensive memory buffers and elaborate architectural designs. Inspired by the potential of prompt-based strategies, our study adopts pre-trained models as foundational structures, capitalizing on prompt tokens to facilitate continual learning in scenarios plagued by missing modalities.

\section{Method}
\begin{figure}[t] \label{fig:arch}
\begin{center}
    \includegraphics[width=0.9\linewidth]{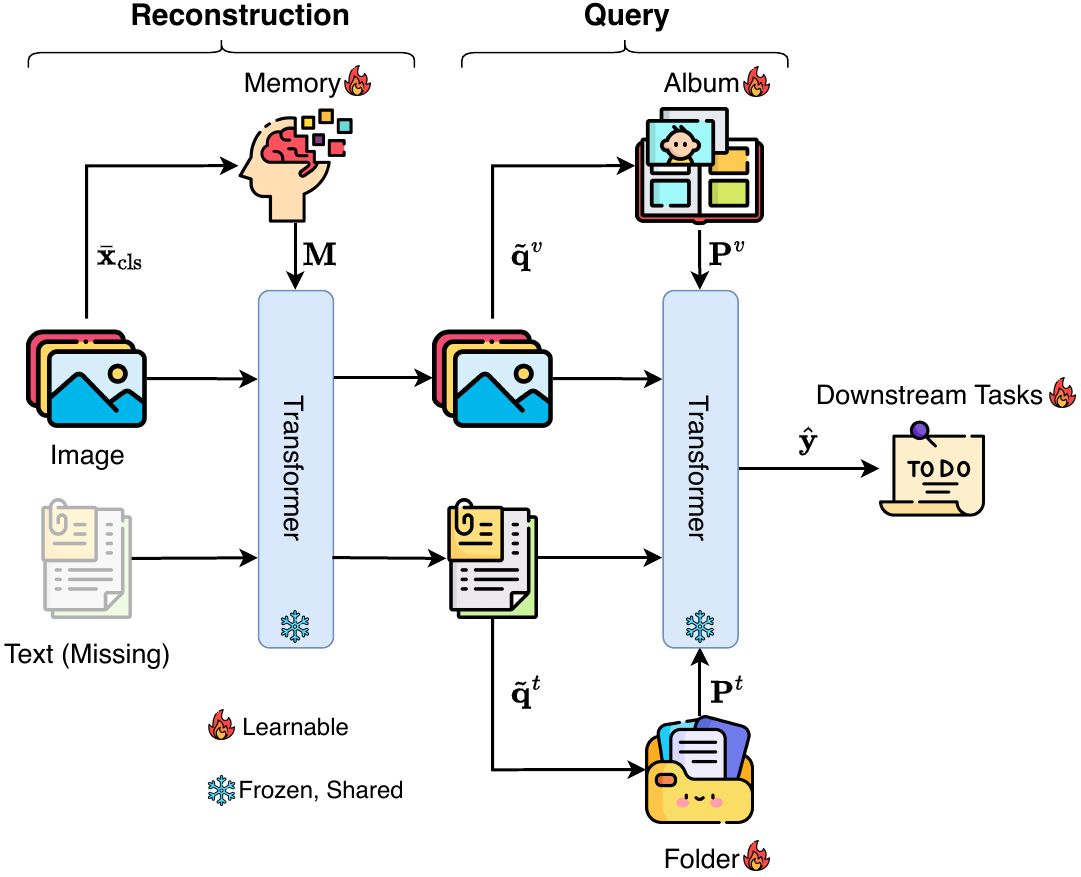}
\end{center}
\caption{Pipeline of the {\shortmethod} framework~(Text modality is unavailable in the figure). The parameters of LMMs are frozen, and introducing prompt learning enables parameter-efficient fine-tuning on user devices. When meeting the missing modality, the available modality is utilized to generate memory prompts from a \texttt{Memory} pool via a key-query mechanism, and memory prompts are inserted into LMMs to reconstruct the missing modality as the modality-specific query. Then, the two modality-specific queries are utilized to generate modality-specific prompts from a text pool~(\texttt{Folder}) and a visual pool~(\texttt{Album}), which are used to learn downstream tasks.}
\end{figure}

\subsection{Problem Definition}
In this paper, we focus on multi-modal learning with missing modalities during continual fine-tuning. Additionally, it is crucial to develop the method without the need to fine-tune the entire pre-trained model due to the limited resources in user devices. To simplify but without loss of generality, we consider a multi-modal dataset with text (t) and visual (v) modalities. Given  a specific task, we denote its training datasets by  $\mathcal{D}$ is built upon three subsets:
\begin{equation}
    \mathcal{D} = \{ \mathcal{D}^{\mathrm{t}}, \mathcal{D}^{\mathrm{v}}, \mathcal{D}^{\mathrm{c}}\}.
\end{equation}
Specifically, the subset $\mathcal{D}^{\mathrm{t}} = \{(\mathbf{t}_{i}, \mathbf{y}_{i})\}_{i=1}^L$ contains modality-incomplete data samples with text only, $\mathcal{D}^{\mathrm{v}} = \{(\mathbf{v}_{i}, \mathbf{y}_{i})\}_{i=1}^M$ contains visual~(image)-only data samples, and $\mathcal{D}^{\mathrm{c}} = \{(\mathbf{t}_{i}, \mathbf{v}_{i}, \mathbf{y}_{i})\}_{i=1}^N$ contains modality-complete samples with both text and image, where $\mathbf{t}_i$ denotes the embedding of textual content and $\mathbf{v}_i$ corresponds to the embedding of visual content, and $\mathbf{y}_i \in \mathbb{R}^C $ is the label vector where $C$ is the number of classes.  Following~\cite{DBLP:conf/cvpr/LeeTCL23}, we assign dummy inputs $\tilde{\mathbf{t}}/\tilde{\mathbf{v}}$ to represent missing-modality data and obtain $\tilde{\mathcal{D}}^{\mathrm{t}}= \{(\mathbf{t}_{i}, \tilde{\mathbf{v}}, \mathbf{y}_{i})\}_{i=1}^L$ and $ \tilde{\mathcal{D}}^{\mathrm{v}} = \{(\tilde{\mathbf{t}}, \mathbf{v}_{i}, \mathbf{y}_{i})\}_{i=1}^M$.  During the training, data from previous tasks is not available. In the inference phase, the task identity is unknown.

\subsection{Modality-Specific Prompt Learning}

User devices only have limited resources, which cannot re-train or fine-tune the entire models with numerous parameters. Thus, we turn our attention to the prompt-based continual learning methods~\cite{DBLP:conf/cvpr/0002ZL0SRSPDP22,DBLP:conf/eccv/0002ZESZLRSPDP22,DBLP:conf/cvpr/SmithKGCKAPFK23}, which retrieve prompts from a learnable prompt pool by using an image-wise prompt query while keeping other parameters frozen to maintain the previous knowledge and learn new skills, significantly reducing the resource requirements. 

However, the unimodal prompt pool suffers from capturing knowledge within each modality and entanglement between different modalities~(shown in Section~\ref{sec:effectiveness}). Therefore, for each modality, we create a modality-specific pool for storing the decomposed modality-specific knowledge. Specifically, we create two modality-specific prompt pools, \texttt{Folder} for texts and \texttt{Album} for images. Each modality-specific prompt pool contains learnable attention weights $\mathbf{A}^{m}  \in \mathbb{R}^{D \times K}$, keys $\mathbf{K}^{m} \in \mathbb{R}^{D \times K}$, where $D$ is the dimension, $K$ is the size of the pool, and $m$ denotes a specific modality, which can be the text modality (t) for \texttt{Folder} or the visual modality (v) for \texttt{Album}, \emph{i.e.}, $m \in \{\mathrm{t}, \mathrm{v}\}$. In the meanwhile, we denote the prompt pool for each specific modality by $\mathcal{P}^{m} = \{ \mathbf{P}^{m}_k \}_{k=1}^K$, where $\mathbf{P}^{m}_k \in \mathbb{R}^{D \times N_{\mathrm{p}}} $ and $N_{\mathrm{p}}$ is the length of prompts.

We utilized prompt selection~\cite{DBLP:conf/cvpr/SmithKGCKAPFK23} for obtaining prompts, which consists of three steps: 1) query generation, 2) weight computation, and 3) prompt aggregation. Below we illustrate these three steps, respectively.

\vspace{1mm} 
\noindent \textbf{Query generation.} We convert the text input $\mathbf{t}$ into a sequence of token embedding  $\mathbf{X}^{\mathrm{t}} =  [\mathbf{x}^{\mathrm{t}}_1, \cdots, \mathbf{x}^{\mathrm{t}}_P ]$ and crop the image $\tilde{\mathbf{v}}$ into a sequence of patch embedding $\mathbf{X}^{\mathrm{v}} = [\mathbf{x}_1^{\mathrm{v}}, \cdots, \mathbf{x}_Q^{\mathrm{v}}]$. We concatenate the $\mathbf{X}^{\mathrm{t}}$ and $\mathbf{X}^{\mathrm{v}}$ as well as  two special tokens $\mathbf{x}_{\mathrm{cls}}^{\mathrm{t}}$ and $\mathbf{x}_{\mathrm{cls}}^{\mathrm{v}}$ into a long sequence as the input of an LMM, and generate an output sequence of the same length:
\begin{equation}
\label{eq:qe}
    [\tilde{\mathbf{x}}_{\mathrm{cls}}^{\mathrm{t}}, \tilde{\mathbf{X}}^{\mathrm{t}}, \tilde{\mathbf{x}}_{\mathrm{cls}}^{\mathrm{v}}, \tilde{\mathbf{X}}^{\mathrm{v}}] = \mathrm{LMM}([\mathbf{x}_{\mathrm{cls}}^{\mathrm{t}}, \mathbf{X}^{\mathrm{t}}, \mathbf{x}_{\mathrm{cls}}^{\mathrm{v}},\mathbf{X}^{\mathrm{v}} ]).
\end{equation}
We set $\tilde{\mathbf{x}}_{\mathrm{cls}}^{\mathrm{t}}$ as $\mathbf{q}^{\mathrm{t}}$, which is the query embedding for text  and  $\tilde{\mathbf{x}}_{\mathrm{cls}}^{\mathrm{v}}$ as $\mathbf{q}^{\mathrm{v}}$, which is the query embedding for image.

\vspace{1mm} 
\noindent \textbf{Weight computation.} The generated text/image query embedding is further utilized to generate the weight vector $\mathbf{w}^{m} = [{w}^{m}_1, \cdots, {w}^{m}_K]$ ($m \in \{\mathrm{t}, \mathrm{v}\}$) as:

\begin{equation} \label{eq:weight-vector}
    \begin{aligned}
        w^{\mathrm{t}}_k = {s}( \mathbf{q}^{\mathrm{t}} \odot \mathbf{A}^{\mathrm{t}}_k, \mathbf{K}^{\mathrm{t}}_k), \forall k \in [1,K],  \\
        w^{\mathrm{v}}_k = s(\mathbf{q}^{\mathrm{v}} \odot \mathbf{A}^{\mathrm{v}}_k, \mathbf{K}^{\mathrm{v}}_k),  \forall k \in [1,K],  \\
    \end{aligned}
\end{equation}
where $s(\cdot,\cdot)$ measures consine similarity; $\odot$ is Hadamard product;   $\mathbf{A}^{\mathrm{t}}_k/\mathbf{A}^{\mathrm{v}}_k$ denotes the $k$-th column of $\mathbf{A}^{\mathrm{t}}/\mathbf{A}^{\mathrm{v}}$. 

\vspace{1mm} 
\noindent \textbf{Prompt Aggreation.} Finally, we adopt the computed weights  $\mathbf{w}^{\mathrm{t}} / \mathbf{w}^{\mathrm{v}}  $ in Equation~\eqref{eq:weight-vector} to generate modality-specific prompts by weighted summation over the prompt components:
\begin{equation} \label{eq:weight-summation}
    \begin{aligned}
        \mathbf{P}^{m} = \sum_{k=1}^K w^{m}_k \cdot \mathbf{P}^{m}_k, \forall m \in \{\mathrm{t}, \mathrm{v}\}.
    \end{aligned}
\end{equation}
The generated modality-specific prompts $\mathbf{P}^{\mathrm{t}}$ and $\mathbf{P}^{\mathrm{v}}$ will collaborate in a pre-trained LMM for facilitating downstream multi-modality understanding tasks. We term this process as multi-modality prompt collaboration,  which we will introduce the details in Section~\ref{sec:mmc}.

Note that, the modality-specific prompt learning relies on condition that a sample contains both textual content $\mathbf{t}$  and visual content $\mathbf{v}$ for generating text query embedding $\mathbf{q}^{\mathrm{t}}$ and visual query embedding  $\mathbf{q}^{\mathrm{v}}$.
Nevertheless, as above mentioned, in real scenarios, we might have only access to the content of a single modality and fail to explicitly obtain both  $\mathbf{q}^{\mathrm{t}}$ and $\mathbf{q}^{\mathrm{v}}$. To address the modality-missing challenge, we proposed to reconstruct the query embedding of the missing modality based on a memory bank $\mathcal{M}$.
We will introduce the detailed process of reconstructing the missing query embedding in Section~\ref{sec:mqr}.

\subsection{Missing Query Reconstruction}
\label{sec:mqr}

We create a memory pool $\mathcal{M} = \{ \mathbf{M}_i \}_{i=1}^R$ containing multi-modal reconstruction prompts and harness the knowledge in a pre-trained large multi-modal model~(LMM) built upon a stack of Transformer blocks to reconstruct the missing-modality query $\mathbf{q}^{\mathrm{t}}$ or $\mathbf{q}^{\mathrm{v}}$. 

Let's consider a sample with only text modality available $(\mathbf{t}, \tilde{\mathbf{v}}, \mathbf{y})$
where $\mathbf{t}$ denotes the text, $\tilde{\mathbf{v}}$ is a dummy image. We convert the text input $\mathbf{t}$ into a sequence of token embedding  $\mathbf{X}^{\mathrm{t}} =  [\mathbf{x}^{\mathrm{t}}_1, \cdots, \mathbf{x}^{\mathrm{t}}_P ]$ and crop the image $\tilde{\mathbf{v}}$ into a sequence of patch embedding $\tilde{\mathbf{X}}^{\mathrm{v}} = [\tilde{\mathbf{x}}_1^{\mathrm{v}}, \cdots, \tilde{\mathbf{x}}_Q^{\mathrm{v}}]$. Then, we concatenate the $\mathbf{X}^{\mathrm{t}}$ and $\mathbf{X}^{\mathrm{v}}$ as well as 
a special token $\mathbf{x}_{\mathrm{cls}}$ into a long sequence as the input of an LMM, and generate an output sequence of the same length:

\begin{equation}
\label{constrv}
    [\bar{\mathbf{x}}_{\mathrm{cls}}, \bar{\mathbf{X}}^{\mathrm{t}}, \bar{\mathbf{X}}^{\mathrm{v}} ] = \mathrm{LMM} ([\mathbf{x}_{\mathrm{cls}}, \mathbf{X}^{\mathrm{t}}, \tilde{\mathbf{X}}^{\mathrm{v}} ]).
\end{equation}
We use $\bar{\mathbf{x}}_{\mathrm{cls}}$ as the query and conduct prompt selection based on the memory template pool $\mathcal{M} = \{\mathbf{M}_i\}_{i=1}^M$ and generate the prompts $\mathbf{M} = [\mathbf{m}_1, \cdots, \mathbf{m}_{N_{\mathrm{p}}}]$ using the same prompt selection process described in Equation~\eqref{eq:qe} ~\eqref{eq:weight-vector} ~\eqref{eq:weight-summation}.

We further concatenate the token embeddings $\mathbf{X}^{\mathrm{t}}$, the patch embeddings $\tilde{\mathbf{X}}^{\mathrm{v}}$, the prompts  $\mathbf{M}$ from the memory pool and the special tokens $\mathbf{x}_{\mathrm{cls}}$ into a long sequence as the input of the same LMM used in Equation~\eqref{constrv}:

\begin{equation} \label{eq:mem_rec}
[ \hat{\mathbf{x}}_{\mathrm{cls}}, \hat{\mathbf{M}}, \hat{\mathbf{X}}^{\mathrm{t}}, \hat{\mathbf{X}}^{\mathrm{v}} ] =    \mathrm{LMM}([\mathbf{x}_{\mathrm{cls}}, \mathbf{M}, \mathbf{X}^{\mathrm{t}}, \tilde{\mathbf{X}}^{\mathrm{v}} ]).
\end{equation}
We set $\hat{\mathbf{x}}_{\mathrm{cls}}$ in the output as reconstructed visual query embedding for the downstream task:
\begin{equation} \label{eq:mem_query}
    \hat{\mathbf{q}}^{\mathrm{v}} = \hat{\mathbf{x}}_{\mathrm{cls}}.
\end{equation}
In a similar manner, given a sample with only visual modality available $(\tilde{\mathbf{t}}, {\mathbf{v}}, \mathbf{y})$
where  $\tilde{\mathbf{t}}$ is a  dummy text, we can reconstruct the text query embedding $\hat{\mathbf{q}}^{\mathrm{t}}$.

\begin{minipage}[t]{.45\textwidth}
    \centering
    \begin{algorithm}[H]
\caption{Missing Query Reconstruction}\label{alg:rec}
\textbf{Input}: Modality-incomplete data $({\mathbf{t}}, \tilde{\mathbf{v}}, 
\mathbf{y})$; Memory pool $\mathcal{M}$; Large multi-modal model (LMM)\\
\textbf{Output}: Reconstructed query $\hat{\mathbf{q}}^{\mathrm{v}} $
\begin{algorithmic}[1]
\Statex \# Generate memory query embedding for modality-incomplete data, described in Equation~\eqref{constrv}
\State $\bar{\mathbf{x}}_{\mathrm{cls}} = \mathrm{LMM}({\mathbf{t}}, \tilde{\mathbf{v}})$
\Statex \# Select prompts from the memory pool, described in Equation~\eqref{eq:weight-vector}~\eqref{eq:weight-summation}
\State $\mathbf{M} = \operatorname{select\_prompt}(\bar{\mathbf{x}}_{\mathrm{cls}}, \mathcal{M})$
\Statex \# Reconstruct the missing query with memory prompts, described in Equation~\eqref{eq:mem_rec}~\eqref{eq:mem_query}
\State $\hat{\mathbf{q}}^m = \mathrm{LMM}( \mathbf{t}, \tilde{\mathbf{v}}, \mathbf{M})$ 
\end{algorithmic}
\end{algorithm}
\end{minipage}
\hfill
\begin{minipage}[t]{.45\textwidth}
   \centering
    \begin{algorithm}[H]
\caption{RebQ Pipeline}\label{alg:cls}
\textbf{Input}: Modality-incomplete data $({\mathbf{t}}, \tilde{\mathbf{v}}, 
\mathbf{y})$; Memory prompt pool $\mathcal{M}$; Text prompt pool $\mathcal{P}^{\mathrm{t}}$; Image prompt pool $\mathcal{P}^{\mathrm{v}}$; Large multi-modal model (LMM); 
\\
\textbf{Output}: Predicted outputs $\hat{\mathbf{y}}$
\begin{algorithmic}[1]

\Statex \# Generate query embedding for the available text modality, described in Equation~\eqref{eq:qe}
\State $\mathbf{q}^{\mathrm{t}}, \bar{\mathbf{x}}_{\mathrm{cls}} = \mathrm{LMM}(\mathbf{t}, \tilde{\mathbf{v}})$
\Statex \# Reconstruct missing vision query embeddings, described in Algorithm~\ref{alg:rec}
\State $\hat{\mathbf{q}}^{\mathrm{v}} = \mathrm{reconstruct}( \textbf{t},  \bar{\mathbf{x}}_{\mathrm{cls}}, \mathcal{M})$
\Statex  \# Select prompts from the pool, described in Equation~\eqref{eq:weight-vector}~\eqref{eq:weight-summation}
\State $\mathbf{P}^{\mathrm{t}} = \operatorname{select\_prompt}(\mathbf{q}^{\mathrm{t}}, \mathcal{P}^{\mathrm{t}})$
\State $\mathbf{P}^{\mathrm{v}} = \operatorname{select\_prompt}(\hat{\mathbf{q}}^{\mathrm{v}}, \mathcal{P}^{\mathrm{v}})$

\Statex \# Predict outputs for downstream tasks, described in Equation~\eqref{eq:output}
\State $\hat{\mathbf{y}} = \mathrm{LMM}(\mathbf{t}, \tilde{\mathbf{v}}, \mathbf{P}^{\mathrm{t}}, \mathbf{P}^{\mathrm{v}})$

\end{algorithmic}
\end{algorithm}
\end{minipage}

For each modality-complete sample, $(\mathbf{t}_{i}, \mathbf{v}_{i}, \mathbf{y}_{i})$, we construct its missing-modality counterparts by replacing the original embedding $\mathbf{t}_{i}/\mathbf{v}_{i}$ by the dummy input $\tilde{\mathbf{t}}_{i}/\tilde{\mathbf{v}}_{i}$ to obtain the visual-only sample $(\tilde{\mathbf{t}}_{i}, {\mathbf{v}}_{i}, \mathbf{y}_{i})$ and the text-only sample $(\mathbf{t}_{i}, \tilde{\mathbf{v}}_{i}, \mathbf{y}_{i})$.
After that, we use the miss-modality samples to generate the reconstructed text query embedding $\hat{\mathbf{q}}^{\mathrm{t}}_i$  and the reconstructed visual query embedding $\hat{\mathbf{q}}^{\mathrm{v}}_i$.
In the meanwhile, we obtain the ground-truth text/visual query embedding $\mathbf{q}^{\mathrm{t}}/\mathbf{q}^{\mathrm{v}}$ based on the original text/visual input as Equation~\eqref{eq:qe}. 
We compute the reconstruction loss to learn the prompts for assisting reconstruction:

\begin{equation}
    \begin{aligned}
    \mathcal{L}_r = \frac{1}{N} \Big \{ \sum_{i=1}^N \| \mathbf{q}^{\mathrm{t}}_i - \hat{\mathbf{q}}^{\mathrm{t}}_i \|_2^2 + \| \mathbf{q}^{\mathrm{v}}_i  - \hat{\mathbf{q}}^{\mathrm{v}}_i \|_2^2  \Big \}.
    \end{aligned}
\end{equation}

\subsection{Multi-modality Prompt Collaboration}
\label{sec:mmc}

The textual/visual content $\mathbf{t}$/$\mathbf{v}$ and the modality-specific prompts $\mathbf{P}^{\mathrm{t}}$/$\mathbf{P}^{\mathrm{v}}$ obtained from Equation~\eqref{eq:weight-summation} serve as the input of the large multi-modal model (LMM) for target multi-modality understanding task:
\begin{equation} \label{eq:output}
    \hat{\mathbf{y}} = \mathrm{LMM}(\mathbf{t}, \mathbf{v}, \mathbf{P}^{\mathrm{t}}, \mathbf{P}^{\mathrm{v}}),
\end{equation}
where $\hat{\mathbf{y}} \in \mathbb{R}^C $ is the predicted logits and $C$ denotes the number of classes. Specifically, to obtain $\hat{\mathbf{y}}$, we first obtain the embedding of cls token in the output of the LMM, $\mathbf{x}^{\mathrm{out}}_{\mathrm{cls}}$. After that, we project  cls token embedding into the predicted logits  $\hat{\mathbf{y}}$ by a head $h(\cdot)$ with a  fully-connect layer:
\begin{equation}
    \hat{\mathbf{y}} = h(\mathbf{x}^{\mathrm{out}}_{\mathrm{cls}})
\end{equation}

In this work, 
we address the classification task and adopt cross-entropy~(CE) loss~(binary cross-entropy loss for the multi-label classification task) for training:

\begin{equation}
    \mathcal{L}_{\mathrm{c}} = \mathrm{CE}(\hat{\mathbf{y}}, \mathbf{y}),
\end{equation}
where $\mathbf{y}$ denotes the groundtruth class label vector.
In the training process, we freeze the weights of the LMM and learn the memory template pool $\mathcal{M}$, text/image prompt pools $\mathcal{P}^{\mathrm{t}}/\mathcal{P}^{\mathrm{v}}$, and the weights of the prediction head $h(\cdot)$. The objective is jointly optimized on two loss functions:
\begin{equation} \label{eq:train}
    \begin{aligned}
       \mathcal{L} = \mathcal{L}_c + \lambda\mathcal{L}_r,
    \end{aligned}
\end{equation}
where $\lambda$ is a hyper-parameter to balance the two loss values.

The algorithms of reconstruction stage and {\shortmethod} are listed in Algorithm~\ref{alg:rec} and Algorithm~\ref{alg:cls}, respectively.

\section{Experiments} \label{sec:exp}
\textbf{Datasets and {\shorttask} Setting. } We construct two curated datasets to facilitate the {\shorttask} task: {\datasetone} and {\datasettwo}. {\datasetone}, derived from the UPMC-Food101~\cite{DBLP:conf/icmcs/WangKTCP15} dataset, containing $101$ food categories and $61,142$ training, $6,846$ validation, and $22,716$ test image-text pairs. We eliminated the category with the fewest samples and allocated the categories into $10$ distinct sessions. The constructed {\datasettwo} dataset is based on the MM-IMDb~\cite{DBLP:conf/iclr/OvalleSMG17} dataset. It is a multi-label classification dataset across $27$ distinct movie genres, consisting of $15,552$ training, $2,608$ validation, and $7,799$ test image-text pairs. Adhering to the multi-label continual learning paradigm~\cite{dong2023knowledge}, we refine the dataset by excluding $7$ most infrequently represented categories, subsequently organizing the remaining into $5$ sessions, each incorporating $4$ categories. 
To set up the missing modality, we focus on a more challenging task wherein the absence of modality can occur in both the training and inference stages. Following~\cite{DBLP:conf/cvpr/LeeTCL23}, we quantify the missing ratio as $\eta\%$, representing the ratio of data with incomplete modalities within each session. For scenarios where only one modality is missing, the proportion of modal-incomplete to modal-complete data adheres to a ratio of $\eta\%$ to $1-\eta\%$. Besides, in instances where both image and text modalities are missing, the dataset contains $\frac{\eta}{2}\%$ image-only data and $\frac{\eta}{2}\%$ text-only data, complemented with $1-\eta\%$ of data retaining both modalities. This configuration highlights the challenges posed by modality scarcity and underscores the robustness in continual missing modality learning environments.

\noindent\textbf{Comparison Methods.} Because our work introduces a novel task, it presents unique challenges in direct comparison with existing methods. We have strived to ensure a comprehensive and fair evaluation of our approach. From the missing modality perspective, MAP~\cite{DBLP:conf/cvpr/LeeTCL23} and MSP~\cite{DBLP:journals/corr/abs-2312-15890} are selected because they focus on leveraging pre-trained LMMs to solve the image and text missing modality during both the training and testing stages. From the continual learning perspective, considering that our approach uses prompt learning to harness the rich knowledge in pre-trained LMMs, we choose two prompt-based continual learning algorithms, L2P~\cite{DBLP:conf/cvpr/0002ZL0SRSPDP22} and DualPrompt~\cite{DBLP:conf/eccv/0002ZESZLRSPDP22}, as comparisons.

\begin{table}[t]
    \caption{Results on {\datasetone}. The evaluation metrics, AP and FG, are based on the classification accuracy. Missing modality occurs in both training and testing.}
    \label{tab:food101}
    \begin{center}
    \resizebox{0.9\linewidth}{!}{
    \begin{tabular}{@{}ccc|rr|rr|rr|rr|rr@{}}
    \toprule
    \multirow{2}{*}{$\eta$} & \multirow{2}{*}{\#Image} & \multirow{2}{*}{\#Text} & \multicolumn{2}{c|}{MAP~\cite{DBLP:conf/cvpr/LeeTCL23}} & \multicolumn{2}{c|}{MSP~\cite{DBLP:journals/corr/abs-2312-15890}} & \multicolumn{2}{c|}{L2P~\cite{DBLP:conf/cvpr/0002ZL0SRSPDP22}} & \multicolumn{2}{c|}{DualPrompt~\cite{DBLP:conf/eccv/0002ZESZLRSPDP22}} & \multicolumn{2}{c}{RebQ} \\
     &  &  & AP~($\uparrow$) & FG~($\downarrow$) & AP~($\uparrow$) & FG~($\downarrow$) & AP~($\uparrow$) & FG~($\downarrow$)  & AP~($\uparrow$) & FG~($\downarrow$) & AP~($\uparrow$) & FG~($\downarrow$) \\ \midrule
    \multirow{3}{*}{$10\%$} & $100\%$ & $90\%$  & 20.66 & 82.50 & 21.45 & 80.03 & 34.09 & 5.80 & 59.56 & 3.38 & \textbf{68.67} & 9.50  \\ 
                             & $90\%$ & $100\%$ & 21.53 & 82.20 & 23.29 & 78.98 & 35.21 & 5.51 & 59.90 & 1.72 &  \textbf{72.46} & 7.64  \\ 
                             & $95\%$ & $95\%$  & 22.84 & 80.13 & 22.12 & 79.35 & 34.00 & 6.55 & 58.18 & 6.32 &  \textbf{71.06} & 8.22  \\ \midrule
    \multirow{3}{*}{$30\%$} & $100\%$ & $70\%$ & 16.57 & 83.77 & 21.37 & 79.95 & 30.40 & 6.53 & 51.86 & 6.46 &  \textbf{62.06} & 10.76  \\ 
                             & $70\%$ & $100\%$ & 24.00 & 77.12 & 22.21 & 76.02 & 34.75 & 4.66 & 52.66 & 5.11 & \textbf{71.62} & 6.44  \\ 
                             & $85\%$ & $85\%$  & 20.66 & 79.81 & 21.89 & 79.01 & 29.76 & 6.87 & 48.56 & 8.73 & \textbf{66.37} & 8.07  \\ \midrule
    \multirow{3}{*}{$50\%$} & $100\%$ & $50\%$ & 18.18 & 79.77 & 17.88 & 77.31 & 28.95 & 6.33 & 47.70 & 6.09 &  \textbf{55.87} & 12.07  \\ 
                             & $50\%$ & $100\%$ & 23.85 & 75.59 & 21.57 & 77.99 & 32.18 & 4.55 & 50.22 & 5.03 & \textbf{69.23} & 5.84  \\ 
                             & $75\%$ & $75\%$  & 18.66 & 79.80 & 18.10 & 78.04 & 25.30 & 6.17 & 43.67 & 8.52 & \textbf{62.40} & 8.24 \\ \midrule
    \multirow{3}{*}{$70\%$} & $100\%$ & $30\%$  & 17.68 & 77.82 & 19.29 & 78.17 & 26.57 & 5.80 & 43.09 & 8.96 &  \textbf{50.00} & 12.47  \\ 
                             & $30\%$ & $100\%$ & 22.48 & 75.21 & 21.22 & 75.21 & 30.43 & 3.99 & 50.28 & 5.19 & \textbf{69.41} & 3.73  \\ 
                             & $65\%$ & $65\%$  & 20.00 & 75.95 & 19.76 & 76.54 & 24.62 & 5.66 & 40.69 & 7.26 & \textbf{59.92} & 8.56 \\ \midrule
    \multirow{3}{*}{$90\%$} & $100\%$ & $10\%$  & 16.92 & 76.60 & 18.62 & 75.14 & 24.94 & 5.60 & 37.26 & 11.46 & \textbf{48.15} & 12.76  \\ 
                             & $10\%$ & $100\%$ & 24.89 & 70.84 & 21.69 & 71.23 & 29.77 & 4.54 & 51.16 & 4.80 &  \textbf{67.71} & 4.71 \\ 
                             & $55\%$ & $55\%$ & 18.41 & 75.19 & 20.45 & 74.31 & 24.62 & 5.66 & 40.69 & 7.29 & \textbf{54.67} & 8.78  \\
    \bottomrule
    \end{tabular}
    }
    \end{center}
\end{table}

\vspace{1mm} 
\noindent\textbf{Evaluation Metrics.} In the {\shorttask} setting, we employ two standard continual learning metrics~\cite{DBLP:conf/eccv/ChaudhryDAT18,DBLP:conf/iclr/ChaudhryRRE19,DBLP:conf/nips/Lopez-PazR17,DBLP:conf/cvpr/ZhangZX23}: Average Performance~(AP) and Average Forgetting~(FG). AP is the average performance calculated from all tasks. Assume $a_{i,j}$ is the testing performance on the $i$-th task after training the model on the $j$-th task, $\mathrm{AP}=\frac{1}{T}\sum_{t=1}^{T}a_{t, T}$. In contrast, FG measures performance degradation in subsequent tasks, $\mathrm{FG} = \frac{1}{T-1}\sum_{t=1}^{T-1}\max_{z\in\{t,\cdots,T-1\}}(a_{t,z}-a_{t,T})$. We compute Accuracy in {\datasetone} and F1-Macro in {\datasettwo}.

\vspace{1mm} 
\noindent\textbf{Implementation Details.} Following~\cite{DBLP:conf/cvpr/LeeTCL23}, we utilize  ViLT\footnote{https://huggingface.co/dandelin/vilt-b32-mlm}~\cite{DBLP:conf/icml/KimSK21} as the pre-trained LMM. An empty string denotes the dummy text input~(i.e., $\tilde{\mathbf{t}}$), and an image with all pixel values equal one as the dummy image input~(i.e., $\tilde{\mathbf{v}}$). All the parameters in the LMM are frozen and a learnable fully-connected layer is added as a classifier head. For memory, image, and text prompt pools, we set the length of prompts, the number of prompt layers, and the prompt pool size to $8$, $8$, and $128$, respectively. All the prompts in pools are randomly initialized and then optimized during training process. The batch size is $4$. We use the AdamW~\cite{DBLP:conf/iclr/LoshchilovH19} optimizer, and the learning rate is set to $1$e-$4$. We adopt the linear warmup cosine annealing scheduler to adjust the learning rate, and $10\%$ of the total training steps are used to warm up. $\lambda$ is set to $0.01$. All experiments are conducted on one NVIDIA A5000 GPU.

\begin{table}[h!]
    \caption{Results on {\datasettwo}. The evaluation metrics, AP and FG, are based on the F1-Macro. Missing modality occurs during both training and testing.}
    \label{tab:mmimdb}
    \begin{center}
    \resizebox{0.9\linewidth}{!}{
    \begin{tabular}{@{}ccc|rr|rr|rr|rr|rr@{}}
    \toprule
    \multirow{2}{*}{$\eta$} & \multirow{2}{*}{\#Image} & \multirow{2}{*}{\#Text} & \multicolumn{2}{c|}{MAP~\cite{DBLP:conf/cvpr/LeeTCL23}} & \multicolumn{2}{c|}{MSP~\cite{DBLP:journals/corr/abs-2312-15890}} & \multicolumn{2}{c|}{L2P~\cite{DBLP:conf/cvpr/0002ZL0SRSPDP22}} & \multicolumn{2}{c|}{DualPrompt~\cite{DBLP:conf/eccv/0002ZESZLRSPDP22}} & \multicolumn{2}{c}{RebQ} \\
     &  &  & AP~($\uparrow$) & FG~($\downarrow$) & AP~($\uparrow$) & FG~($\downarrow$) & AP~($\uparrow$) & FG~($\downarrow$)  & AP~($\uparrow$) & FG~($\downarrow$) & AP~($\uparrow$) & FG~($\downarrow$) \\ \midrule
    \multirow{3}{*}{$10\%$} & $100\%$ & $90\%$  & 15.77 & 34.67 & 19.49 & 35.11 & 13.67 & 8.96 & 22.79 & 20.42 & \textbf{24.19} & 32.46  \\ 
                             & $90\%$ & $100\%$ & 16.63 & 33.30 & 19.01 & 34.89 & 13.54 & 8.26 & 24.82 & 24.09 & \textbf{27.35} & 28.88  \\ 
                             & $95\%$ & $95\%$  & 17.24 & 31.28 & 15.69 & 32.17 & 13.90 & 8.42 & 24.19 & 24.04 & \textbf{26.90} & 28.26  \\ \midrule
    \multirow{3}{*}{$30\%$} & $100\%$ & $70\%$ & 19.85 & 32.14 & 20.01 & 34.87 & 10.68 & 8.03 & 22.96 & 23.96 & \textbf{24.75} & 25.86  \\ 
                             & $70\%$ & $100\%$ & 17.19 & 31.09 & 18.31 & 32.45 & 10.98 & 7.27 & 20.98 & 19.67 & \textbf{26.68} & 27.89  \\ 
                             & $85\%$ & $85\%$  & 14.44 & 33.29 & 14.20 & 35.57 & 11.58 & 7.90 & 22.18 & 20.72 & \textbf{25.80} & 26.95  \\ \midrule
    \multirow{3}{*}{$50\%$} & $100\%$ & $50\%$  & 19.19 & 36.18 & 15.93 & 35.29 & 9.54 & 5.54 & 20.04 & 21.48 & \textbf{23.89} & 25.27  \\ 
                             & $50\%$ & $100\%$ & 18.83 & 32.02 & 16.78 & 34.14 & 9.77 & 4.88 & 19.97 & 20.24 & \textbf{26.45} & 27.53  \\ 
                             & $75\%$ & $75\%$  & 17.36 & 29.67& 16.24 & 31.81 & 10.08 & 5.23 & 20.32 & 19.09 & \textbf{25.31} & 25.88  \\ \midrule
    \multirow{3}{*}{$70\%$} & $100\%$ & $30\%$  & 17.45 & 27.14& 18.01 & 27.32 & 7.26 & 5.37 & 16.80 & 13.82 & \textbf{22.41} & 24.39  \\ 
                             & $30\%$ & $100\%$ & 16.09 & 34.59 & 14.30 & 35.72 & 7.49 & 5.20 & 18.42 & 15.32 & \textbf{26.27} & 27.56  \\ 
                             & $65\%$ & $65\%$  & 17.20 & 29.39 & 18.27 & 29.34 & 6.93 & 6.23 & 16.19 & 15.76 & \textbf{23.77} & 25.26  \\ \midrule
    \multirow{3}{*}{$90\%$} & $100\%$ & $10\%$  & 18.85 & 25.03 & 18.01 & 29.14 & 5.98 & 5.82 & 15.50 & 11.27 & \textbf{21.37} & 23.66  \\ 
                             & $10\%$ & $100\%$ & 17.79 & 31.65 & 17.88 & 32.19 & 6.29 & 5.09 & 21.22 & 18.38 & \textbf{26.22} & 27.50  \\ 
                             & $55\%$ & $55\%$  & 16.28 & 27.89 & 16.20 & 31.47 & 6.84 & 6.39 & 16.82 & 8.44 & \textbf{25.42} & 25.82  \\
    \bottomrule
    \end{tabular}    }
    \end{center}
\end{table}

\begin{figure}[t]
\begin{center}
    \includegraphics[width=0.6\linewidth]{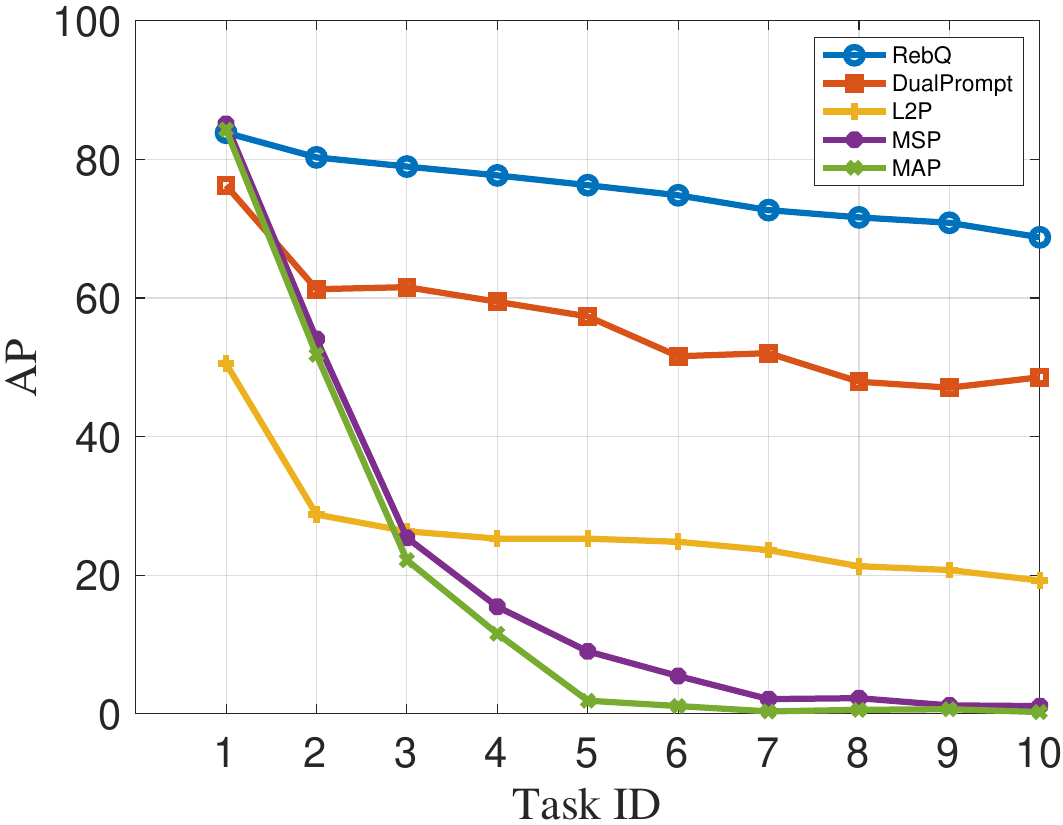}
\end{center} 
\caption{Results of each incremental stage on UPMC-Food101-CMML. $\eta$ is $70\%$.}
\label{fig:incremental-stage}
\end{figure}

\subsection{Main Results}

Table~\ref{tab:food101} summarizes the experimental results of the {\datasetone} dataset, demonstrating that {\shortmethod} consistently surpasses the comparative method across various configurations in both Average Precision~(AP) and Forgetting~(FG). Unlike the MAP and MSP, which employ prompt vectors for distinct missing types~(\textit{i.e.}, text-only or image-only) and encounters significant catastrophic forgetting due to overfitting of prompt vectors on the current task, {\shortmethod} decomposes the unified prompts into modality-specific components stored in pools and accesses the pools via reconstructed missing queries, markedly improving knowledge transfer for subsequent tasks. L2P and DualPrompt also employ the prompt pool, but they cannot access the pool when modality-missing data arrives, leading to lower AP compared with our proposed {\shortmethod}. Note that L2P and DualPrompt achieve lower FG in some cases. The reason is their relatively lower precision, which limits the decreasing range, as shown in Figure~\ref{fig:incremental-stage}. Furthermore, a direct correlation is observed between the quantity of text data and improved performance. We argue that the texts in the {\datasetone} dataset are webpage titles collected from the Internet, which may explicitly include the class label and introduce a strong classification bias.

Table~\ref{tab:mmimdb} illustrates the results for the {\datasettwo} dataset, where {\shortmethod} markedly surpasses the baseline methods, demonstrating consistent superiority even in the challenging multi-label continual missing modality learning setting. Additionally, we also note a significant correlation between the quality of text modality and AP. We argue that the {\datasetone} dataset comprises movie posters as visual inputs and plots as text inputs. The latter contains detailed descriptions, which are beneficial to classify movie genres.

Figure~\ref{fig:incremental-stage} illustrates the results of each incremental stage. The compared methods suffer from catastrophic forgetting, leading to severe performance drops. However, our proposed {\shortmethod} maintains the performance during incremental stages, demonstrating the effectiveness of our method.

\subsection{Ablation Studies} \label{sec:effectiveness}
In this section, we explore various {\shortmethod} variants to assess the importance of each component. Experiments are on the {\datasetone} dataset with a missing ratio $\eta$ of $70\%$, where both image and text modalities are absent.

\begin{table}[t] 
    \caption{Ablation study on the {\datasetone} dataset.}
    \label{tab:rec}
    \begin{center}
    \begin{tabular}{@{}lrr@{}}
    \toprule
    Method & AP~($\uparrow$) & FG~($\downarrow$) \\ \midrule
    {\shortmethod} & \textbf{59.92} & \textbf{8.56} \\ \midrule
    w/o Reconstruction & 39.22 & 16.27 \\ 
    w/o Modality-Specific Query & 51.82 & 11.92 \\
    w/o Memory Pool & 53.36 & 12.04 \\
    w/o Modality-Specific Pool & 52.07 & 12.29 \\
    \bottomrule
    \end{tabular}
    \end{center}
\end{table}
\begin{figure}[t]
\begin{center}
    \includegraphics[width=0.95\linewidth]{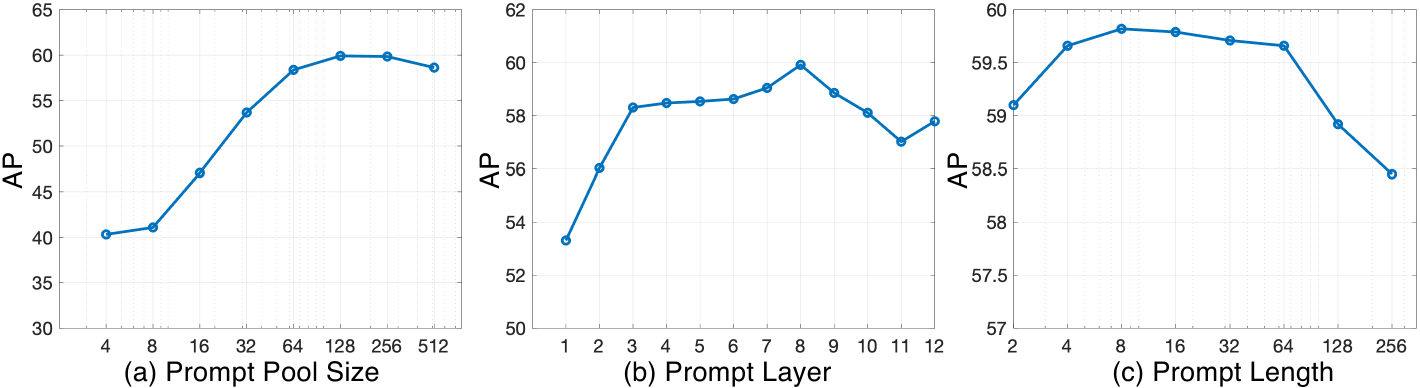}
\end{center} 
\caption{Prompt designs. (a) The pool size of memory/image/text pool. (b) The layers that are inserted prompts. (c) The prompt length.}
\label{fig:prompt-design}
\end{figure}

\begin{figure}[t]
\begin{center}
    \includegraphics[width=0.95\linewidth]{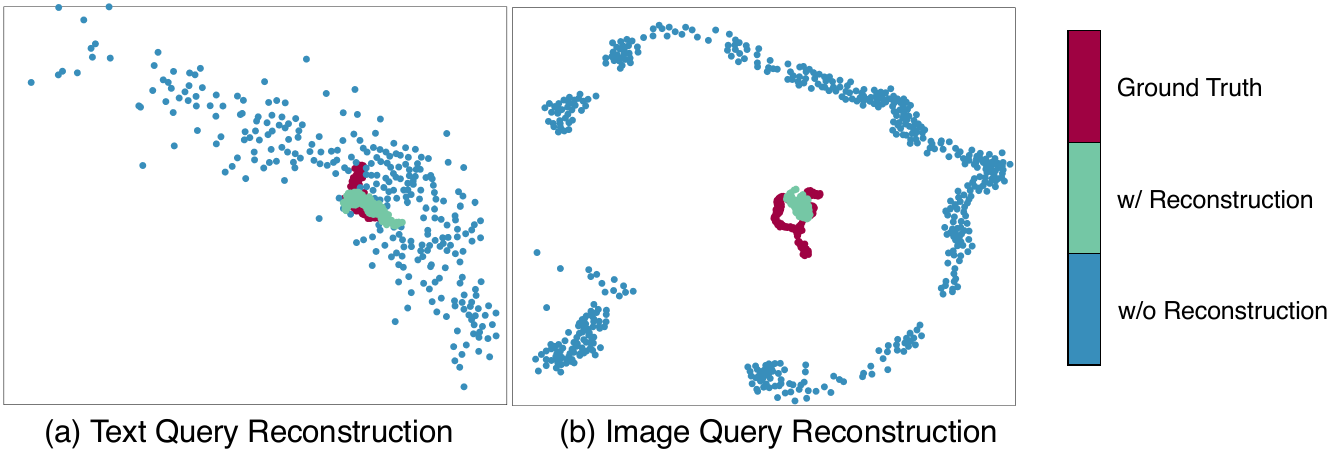}
\end{center}
\caption{t-SNE visualization of Reconstruction. Each point represents a query embedding of dimension $768$.}
\label{fig:tsne}
\end{figure}

\noindent\textbf{Reconstruction.} Vanilla modality-specific prompts assume the data is modality-complete, and prompts can be optimized during training with fully paired data, which cannot handle the CMML setting. We remove the reconstruction stage and directly use the modality-specific queries described in Equation~\eqref{eq:qe} to generate prompts from pools, denoted as ``w/o Reconstruction.'' Table~\ref{tab:rec} illustrates that the performance of ``w/o Reconstruction'' is significantly decreasing due to the query embeddings from the absent modality lacking substantial semantic information, thereby hindering effective prompt generation. To further assess the impact, we introduce a variant ``w/o Modality-Specific Query'' to benchmark the performance using only the available modalities. Specifically, if only one modality is available, we utilize it to generate the single-modality prompts and insert them into the pre-trained LMM. The deteriorating results of the ``Unimodal Query'' underscore the effectiveness of multi-modality prompt collaboration. Besides, we argue that missing modalities can potentially be reconstructed through prompting in pre-trained LMMs. However, it assumes a consistent data distribution, a condition not met in continuous fine-tuning settings. Thus, we develop ``w/o Memory Pool,'' relying on a singular memory prompt vector instead of a memory pool. This approach, as the results indicate, leads to catastrophic forgetting. Finally, we employ t-SNE~\cite{van2008visualizing} to analyze the query distribution. By removing one modality from a modality-complete dataset $\mathcal{D}^{\mathrm{c}}$, we generated two modality-incomplete datasets $\mathcal{D}^{\mathrm{t}}$ and $\mathcal{D}^{\mathrm{v}}$. Subsequently, we extracted query embeddings from $\mathcal{D}^{\mathrm{c}}$ directly using the pre-trained LMM as the ``Ground Truth.'' ``w/ Reconstruction'' refers to the extraction of query embeddings from $\mathcal{D}^{\mathrm{t}}$ and $\mathcal{D}^{\mathrm{v}}$ inclusive of the reconstruction stage, while ``w/o Reconstruction'' denotes extraction without the reconstruction stage. As depicted in Figure~\ref{fig:tsne}, after training with only a tiny partial of full modal data, the distribution alignment of ``Ground Truth'' and ``w/ Reconstruction'' is notably closer compared to ``w/o Reconstruction,'' validating the reconstruction effectiveness of {\shortmethod}. The hyper-parameter $\lambda$ is utilized to balance the classification and reconstruction losses. Table~\ref{tab:lambda} shows the results of sensitivity.

\begin{table}[t]
\centering
\begin{minipage}{.48\linewidth}
    \begin{table}[H]
    \caption{Comparison of different ways of prompt insertion. Attention denotes the prompts inserted into the key and value  in MSA layers. Input means the prompts are appended to the input tokens.}
    \label{tab:prompt-position}
    \begin{center}
    \begin{tabular}{@{}ccrrr@{}}
    \toprule
    Image & Text & AP~($\uparrow$) & FG~($\downarrow$)\\ \midrule
    Attention     & Attention    & \textbf{59.92}   &   \textbf{8.56}\\ \midrule
    Attention     & Input    & 41.61   &   3.32\\
    Input     & Attention    & 53.36   &   8.13\\
    Input     & Input    & 37.54   &   2.59\\ \bottomrule
    \end{tabular}
    \end{center}
\end{table}
\end{minipage}\hfill
\begin{minipage}{.48\linewidth}
    \begin{table}[H]
    \caption{Comparison of different ways of prompt generation. Pool denotes the prompts are generated from a prompt pool. Vector means the prompts are a vector.}
    \label{tab:types}
    \begin{center}
    \begin{tabular}{@{}ccrrr@{}}
    \toprule
    Image & Text & AP~($\uparrow$) & FG~($\downarrow$) \\ \midrule
    Pool     & Pool    &  \textbf{59.92}  & \textbf{8.56}\\ \midrule
    Pool     & Vector    & 55.41 & 10.14\\
    Vector     & Pool    & 52.54  & 10.77\\
    Vector     & Vector    & 46.38  & 11.43\\ \bottomrule
    \end{tabular}
    \end{center}
\end{table}
\end{minipage}
\end{table}

\noindent\textbf{Modality-Specific Prompt.} To further assess the effectiveness of the modality-specific prompt pool, we conducted experiments using a single prompt pool, referred to as ``Unimodal Pool'' in Table~\ref{tab:rec}. The results underscore the significance of modality-specific pools in decomposing knowledge and enhancing knowledge transferability for successive fine-tuning tasks. The performance is also influenced by various prompting strategies, including the size of the prompt pool, the layer at which prompts are inserted, and the length of the prompts. Figure~\ref{fig:prompt-design}~(a) compares performances across different pool sizes, revealing that smaller pools are still prone to catastrophic forgetting. Performance improves up to a certain pool size, beyond which further enlargement does not yield improvement. Figure~\ref{fig:prompt-design}~(b) illustrates the performance across different numbers of prompted transformer layers in the LMM. For instance, a value of $8$ indicates that prompts are inserted into the first $8$ layers. The results reveal that integrating prompts into too few layers restricts expressive capacity, whereas incorporating them into an excessive number of layers often leads to overfitting. Figure~\ref{fig:prompt-design}~(c) presents a comparison of prompt lengths, where shorter lengths limit learning capacity, and excessively long lengths result in performance decreasing. Additionally, we conduct experiments to demonstrate that prompt pools, rather than vectors, improve the knowledge transfer transferability, illustrated in Table~\ref{tab:types}. ``Pool'' represents the use of a prompt pool for a modality, whereas ``Vector'' refers to learnable vectors. The outcomes indicate that prompt pools substantially mitigate forgetting and maintain average performance. Previous works~\cite{DBLP:conf/cvpr/LeeTCL23,DBLP:conf/eccv/0002ZESZLRSPDP22} have highlighted the criticality of prompt insertion position. We attach the prompts to key and value layers, denoted as ``Attention.'' Appending prompts into the input tokens is denoted as ``Input.'' The results from Table~\ref{tab:prompt-position} show that the ``Attention'' improves the results in the {\shorttask} setting.  

\begin{table}[t]
\caption{Sensitivity of $\lambda$ on UPMC-Food101-CMML. $\eta$ is $70\%$.}
\begin{center}
\begin{tabular}{@{}lccccc@{}}
\toprule
$\lambda$ & 1e-4 &1e-3 & 1e-2 & 1e-1 & 1 \\ \midrule
AP & 58.91 & 59.37 & 59.84 & \textbf{59.92} & 59.82 \\ \bottomrule
\end{tabular} \label{tab:lambda}
\end{center}
\end{table}

\section{Conclusion}
In this paper, we propose a new task,~{\task}, which aims to address the critical issue of fine-tuning large multi-modal models~(LMMs) on resource-constrained devices with modality-incomplete stream data. Additionally, we devise the~{\method}~({\shortmethod}) framework to decompose the modal-specific knowledge into prompt pools and then enable multi-modal knowledge collaboration to enhance knowledge transferability for subsequent fine-tuning tasks. Moreover, to obtain the missing query for accessing corresponding pools, we explicitly harness the rich multi-modal knowledge from pre-trained models to reconstruct the missing query. Our experiments confirm~{\shortmethod}'s effectiveness in maintaining performance and preventing knowledge degradation when faced with modal-incomplete stream data. Our research not only enhances the adaptability of LMMs in privacy-sensitive and resource-limited environments but also opens avenues for further research into more robust and efficient multi-modal learning techniques.


%
%
\bibliographystyle{splncs04}
\bibliography{main}

\newpage
\appendix
\section{Comparison of Different Backbones}
\begin{table}[t]
    \caption{Comparison of different backbones.}
    \label{tab:backbone}
    \begin{center}
    \begin{tabular}{@{}cccc|rr@{}}
    \toprule
    Backbone & $\eta$ & \#Image & \#Text &  AP~($\uparrow$) & FG~($\downarrow$) \\ \midrule
    \multirow{3}{*}{CLIP~\cite{DBLP:conf/icml/RadfordKHRGASAM21}} & \multirow{3}{*}{$70\%$} & $100\%$ & $30\%$ & 39.69 & 15.26 \\ 
    & & $30\%$ & $100\%$ & 60.44 & 14.08 \\ 
    & & $65\%$ & $65\%$ & 53.06 & 6.53 \\ \midrule
    \multirow{3}{*}{ViLT~\cite{DBLP:conf/icml/KimSK21}} & \multirow{3}{*}{$70\%$} & $100\%$ & $30\%$ & 50.00 & 12.47 \\ 
    & & $30\%$ & $100\%$ & 69.41 & 3.73 \\ 
    & & $65\%$ & $65\%$ & 59.92 & 8.56  \\ 
    \bottomrule
    \end{tabular}
    \end{center}
\end{table}
Our proposed {\shortmethod} can integrate with various backbones. We employ the ViLT~\cite{DBLP:conf/icml/KimSK21} model as the default backbone in Section~\ref{sec:exp}. ViLT is a fusion-based model that fuses the visual and text tokens to obtain the fused token. Furthermore, we also employ the other mainstream vision-language architectures, i.e., the dual-encoder-based CLIP~\cite{DBLP:conf/icml/RadfordKHRGASAM21} model, to offer a broader assessment of {\shortmethod}. CLIP is a dual-encoder model, where one encoder processes text and the other processes images. These encoders transform their respective inputs into a shared embedding space.

However, the dual-encoder CLIP model with different dimensions within each encoder brings challenges to integration in {\shortmethod}. Specifically, during the reconstruction stage, the memory prompts from the memory pool are inserted into the available modality-specific encoder to reconstruct the missing query embedding. However, the fixed dimension memory prompts~(e.g., $512$ dimensions) cannot be inserted and jointly learned with the different dimensions~(e.g., $768$ and $512$ dimensions in the visual and text encoders, respectively). Therefore, we introduce an extra fully connected layer to project the memory prompts to the embedding space of the missing modality. After obtaining the image and text embeddings from corresponding encoders, we add the two embeddings and feed them into a classifier. The CLIP model in our experiment is ViT-B-32\footnote{https://huggingface.co/openai/clip-vit-base-patch32}.

From Table~\ref{tab:backbone}, the CLIP model performs worse than the ViLT model, even if trained on larger datasets. We argue that the CLIP model is trained by the contrastive learning strategy and aligns the vision and language modality to a common embedding space. This type of training paradigm is inferior to fusion-based approaches in scenarios requiring complex multi-modal understanding skills, which matches the conclusions in recent multi-modal works~\cite{DBLP:journals/corr/abs-2310-05473,DBLP:journals/corr/abs-2303-16604}.

\section{Analysis of Reconstruction}
In this section, we will analyze why the pre-trained LMMs can reconstruct the missing information. Assume the visual modality is missing, we model the modality-incomplete conditional probability distribution $\mathrm{Pr}(\tilde{\mathbf{v}} \mid \mathbf{t};\mathcal{K})$:

\begin{equation} \label{eq:approx_visual}
    \begin{aligned}
        \arg\max\mathrm{Pr}(\tilde{\mathbf{v}} \mid \mathbf{t};\mathcal{K}) &= \frac{\mathrm{Pr}(\tilde{\mathbf{v}}, \mathbf{t};\mathcal{K})}{\mathrm{Pr}(\mathbf{t};\mathcal{K})} \\
        &\approx \frac{\mathrm{Pr}(\mathbf{v}, \mathbf{t};\mathcal{K})}{\mathrm{Pr}(\mathbf{t};\mathcal{K})} \\
        &= \mathrm{Pr}(\mathbf{v} \mid \mathbf{t};\mathcal{K}),
    \end{aligned}
\end{equation}
where $\mathcal{K}$ is the multi-modal knowledge in LMMs. We assume the modality-incomplete distribution can be estimated by modality-complete distribution.

Similarly, when the text modality is missing, we can obtain:

\begin{equation} \label{eq:approx_text}
    \begin{aligned}
        \arg\max\mathrm{Pr}(\tilde{\mathbf{t}} \mid \mathbf{v};\mathcal{K}) &= \frac{\mathrm{Pr}(\tilde{\mathbf{t}}, \mathbf{v};\mathcal{K})}{\mathrm{Pr}(\mathbf{v};\mathcal{K})} \\
        &\approx \frac{\mathrm{Pr}(\mathbf{t}, \mathbf{v};\mathcal{K})}{\mathrm{Pr}(\mathbf{v};\mathcal{K})} \\
        &= \mathrm{Pr}(\mathbf{t} \mid \mathbf{v};\mathcal{K}),
    \end{aligned}
\end{equation}

Equation~\eqref{eq:approx_visual} and \eqref{eq:approx_text} indicate that the model can reconstruct the missing modality by conditioning the existing modality and multi-modal knowledge. They assume $\{\mathbf{t}, \mathbf{v}, \mathbf{y}\}$ and $\{\tilde{\mathbf{t}}, \mathbf{v}, \mathbf{y}\}$ are sampled from the same distribution. In the~{\shorttask} task, data from sequential tasks has a different distribution, also suffering from catastrophic forgetting of reconstruction ability. Therefore, we create a memory pool to tackle this issue, and its effectiveness is shown in Section~\ref{sec:effectiveness}.

\section{Limitations}
In this paper, we introduce a new task, {\task}~({\shorttask}), which is commonly encountered in real applications and propose a {\shortmethod} framework to tackle this task. Although we utilize two modalities, including vision and language, to demonstrate the effectiveness of our proposed framework, multi-modality contains more modalities, such as depth and audio. In fact, our framework has the ability to process more modalities. Compared with MAP~\cite{DBLP:conf/cvpr/LeeTCL23}, which needs $N^2-1$~($N$ is the number of modalities) types of prompts to represent missing modalities, {\shortmethod} only requires $N$ types of prompt pools, significantly reducing the needed prompt types. On the other hand, we employ the pre-trained LMM with only two available modalities. To extend the supported modalities, we can re-train LMMs on large-scale datasets with more modalities~\cite{shvetsova2022everything},  integrate more modalities while keeping the trained encoders frozen~\cite{DBLP:conf/cvpr/GirdharELSAJM23,DBLP:journals/corr/abs-2310-01852}, or devise new methods as the future direction.

\end{document}